\documentclass[runningheads]{llncs}
\usepackage[T1]{fontenc}
\usepackage{graphicx,verbatim}
\usepackage{textgreek}
\usepackage{dsfont}  

\usepackage{subcaption}
\usepackage{xcolor} 
\usepackage{graphicx}
\usepackage{comment}
\usepackage{amsmath,amssymb} 
\usepackage{color}
\usepackage{paralist}
\usepackage{stackengine}
\usepackage{adjustbox}
\usepackage{caption} 
\usepackage{multirow}
\usepackage{diagbox}
\usepackage[colorlinks,linkcolor=blue]{hyperref}
\usepackage{bbm}
\usepackage{booktabs}
\usepackage{bm}
\usepackage{bbding}
\usepackage{colortbl} 
\usepackage{hyperref}

\usepackage{xcolor} 
\usepackage{amssymb}
\begin{document}
\title{Robust Incomplete-Modality Alignment for Ophthalmic Disease Grading and Diagnosis via Labeled Optimal Transport}
%

\author{Qinkai Yu \inst{1}, Jianyang Xie \inst{2}, Yitian Zhao\inst{3}, Cheng Chen\inst{4}, Lijun Zhang\inst{5}, Liming Chen\inst{6}, Jun Cheng\inst{7}, Lu Liu\inst{1}, Yalin Zheng \inst{2}, Yanda Meng\inst{1 (}\Envelope\inst{)}}  
\authorrunning{Q.Yu et al.}
\institute{
\textsuperscript{$1$} Computer Science Department, University of Exeter, Exeter, UK.\\
\textsuperscript{$2$} Eye and Vision Sciences Department, University of Liverpool, Liverpool, UK.\\
\textsuperscript{$3$} Ningbo Institute of Materials Technology and Engineering, Chinese Academy of Sciences, Ningbo, China. \\
\textsuperscript{$4$} Department of Electrical and Electronic Engineering, The University of Hong Kong, Hong Kong, China.\\
\textsuperscript{$5$} Key Laboratory of System Software, Institute of Software, Chinese Academy of Sciences, Beijing, China.\\
\textsuperscript{$6$} School of Computer Science and Technology, Dalian University of Technology, Dalian, China.\\
\textsuperscript{$7$} Institute for Infocomm Research, A*STAR, Singapore. \\
\email{Y.M.Meng@exeter.ac.uk} \\
}
\maketitle              
\begin{abstract}
Multimodal ophthalmic imaging-based diagnosis integrates colour fundus imaging with optical coherence tomography (OCT) to provide a comprehensive view of ocular pathologies. 
However, the uneven global distribution of healthcare resources often results in real-world clinical scenarios encountering incomplete multimodal data, which significantly compromises diagnostic accuracy. 
Existing commonly used deep learning pipelines, such as modality imputation and distillation methods, face notable limitations: (1) Imputation methods struggle with accurately reconstructing key lesion features, since OCT lesions are localized, while fundus images vary in style. (2) distillation methods rely heavily on fully paired multimodal training data. To address these challenges, we propose a novel multimodal alignment and fusion framework capable of robustly handling missing modalities in the task of ophthalmic diagnostics. 
By considering the distinctive feature characteristics of OCT and fundus images, we emphasise the alignment of semantic features within the same category and explicitly learn soft matching between modalities, allowing the missing modality to utilize existing modality information, achieving robust cross-modal feature alignment under the missing modality. Specifically, we leverage the Optimal Transport (OT) mechanism for multi-scale modality feature alignment: class-wise alignment through predicted class prototypes and feature-wise alignment via cross-modal shared feature transport. Furthermore, we propose an asymmetric fusion strategy that effectively exploits the distinct characteristics of OCT and fundus modalities. Extensive evaluations on three large-scale ophthalmic multimodal datasets demonstrate our model's superior performance under various modality-incomplete scenarios, achieving state-of-the-art performance in both complete modality and inter-modality incompleteness conditions. The implementation code is available at \url{https://github.com/Qinkaiyu/RIMA}.


\end{abstract}

\section{Introduction}
Retinal fundus imaging and Optical Coherence Tomography (OCT) serve as prevalent two and three dimensional imaging modalities for diagnosing a wide range of ophthalmic conditions. In clinical diagnostics settings, fundus imaging is used to assess the distribution of lesions on the retinal surface, while OCT provides cross-sectional scans for detailed examination of more profound retinal abnormalities \cite{wu2023gamma}. These two imaging modalities complement each other and differ in anatomy appearance and informational depth. For instance, fundus imaging vividly displays vascular structures and various pathological observations on the retinal surface but cannot reflect the difference in retinal layer thickness. Conversely, OCT can objectively evaluate retinal thickness and subretinal lesions but cannot visualize retinal blood vessels or other essential retinal features \cite{de2015review}. Compared to single-modality approaches, multimodal diagnostic strategies provide a more comprehensive and accurate assessment of ophthalmic conditions \cite{wu2023gamma} via integrating fundus and OCTs. However, in real-world clinical settings, challenges such as high equipment costs, time constraints often result in incomplete data from one of the modalities. 

Currently, commonly used methods to address the issue of missing modalities can be categorized into two main types: 1) modality imputation \cite{sun2024similar,yang2024incomplete} and 2) distillation methods \cite{saha2024examining,liu2025Incomplete}. 
modality imputation methods fill in missing modality samples or generate the missing modality by performing various transformations on the existing modality, either at the image level or the representation level. However, the modality imputation mechanism for OCT and fundus imaging is limited by the difficulty of accurately reconstructing and aligning key lesion features in the presence of missing modalities. This is because the OCT data may not contain sufficient semantic lesion information due to the natural gaps between slices (\textit {e.g.,} 6 mm gaps between each slice) \cite{fujimoto2016development}. Differently, fundus images exhibit diverse stylistic features among samples due to varying imaging conditions, equipment differences, and pathological changes. These factors significantly impair the ability of modality imputation methods to accurately generate critical semantic features, leading to severe overfitting problems. 
distillation methods employ distillation strategies \cite{liu2025Incomplete} to ensure that the modality fusion process is not constrained by the missing ones. However, these methods rely heavily on complete multimodal data and struggle to manage incomplete modalities during training. Specifically, these methods customize architectures tailored to specific datasets and assume that a complete dataset is available during training while missing modalities only occurred during testing \cite{wu2024deep}. In the cases where OCT data is limited, these approaches are unable to effectively utilize the scarce OCT information to improve diagnostic performance. Therefore, there is a need to develop diagnostic methods that can fully utilize the multimodal information from OCT and fundus imaging while maintaining robust performance in the face of totally missing modality data.

\begin{figure}[t!]
    \centering
    \includegraphics[width=1\linewidth]{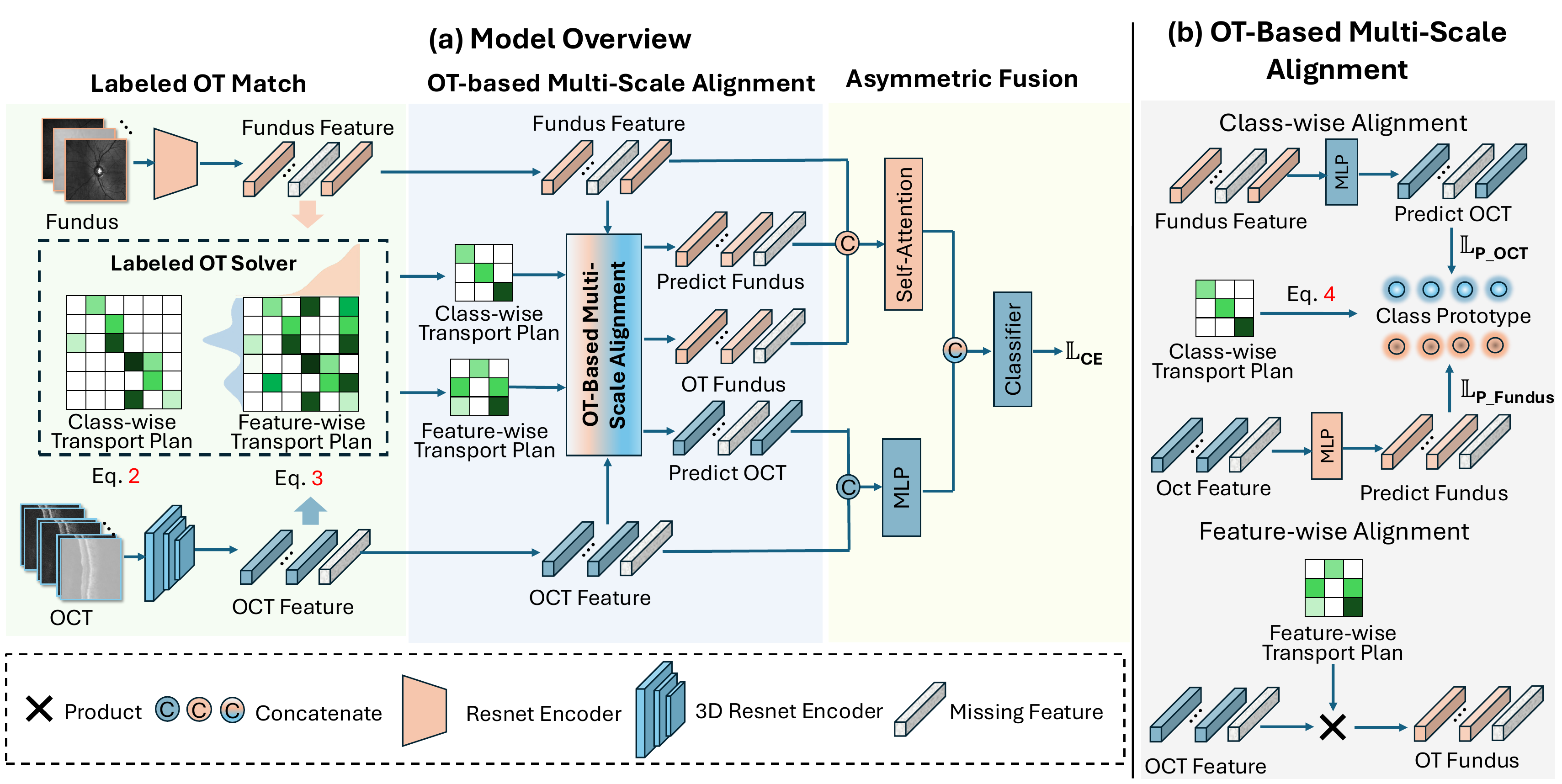}
    \caption{Overview of the proposed framework. (a) The framework consists of three parts: Labeled OT Match, OT-based Multi-Scale Alignment, and Asymmetric Fusion. We leverage the matching results from Optimal Transport to construct class prototypes and cross-modal shared features for multi-scale alignment. To accommodate the distinct characteristics of each modality, we introduce an asymmetric fusion strategy to adaptively integrate modality-specific information. (b) Details of the Multi-Scale Alignment module, showing both class-wise and feature-wise alignment.}
    \label{fig:1}
\end{figure}

In this work, we propose a novel multi-modal alignment and fusion paradigm based on Optimal Transport (OT) for ophthalmic disease diagnosis (overview shown in \autoref{fig:1}) that can efficiently and robustly learn multimodal feature representations in various modality incomplete and complete scenarios. Specifically, OT is employed to achieve class-wise and feature-wise alignment under missing modality conditions, building a bridge between the two modalities so that the missing modality can utilize the information from the existing modality. OT serves as a computational framework for cross-modal alignment, achieving optimal matching with minimal overall cost \cite{duan2022multi,cao2022otkge,wang2021wasserstein,ryu2024cross}. The contributions of this work are summarized as follows: (1) We propose class-wise and feature-wise alignment based on Labeled OT. Different from existing studies \cite{wang2024tuning,xu2023multimodal}, our method neither relies on strict one-to-one alignment nor performs global alignment between the two modalities. Instead, by considering the unique features of OCT and fundus images, we focus the alignment on semantic features within the same class and explicitly learn soft matching between modalities, thereby achieving more robust cross-modal alignment. (2) Additionally, we propose an asymmetric fusion strategy tailored to address the feature discrepancies between OCT and fundus images. (3) We comprehensively evaluate the proposed method on three large-scale ophthalmic multi-modal datasets under various modality-incomplete and complete scenarios. The results demonstrate that our method achieves state-of-the-art performance.

\section{Methods}
\subsection{Labeled OT Match}
Let $\{x,y\} = \{x_{n},y_{n}\}^{N}_{n}$ as a multimodal dataset that has $N$ samples. Each $x_n = \{x_{n}^{f}, x_{n}^{o}\}$ consists of 2 inputs from fundus and OCT modalities, respectively. And $y_n\in{1,2,...,C}$ where $C$ is the number of categories. The modality encoder learn from the input modality $x^{f}$ and $x^{o}$ to produce the single-modal representation $e^f \in \mathbb{R}^{N\times D_{f}}$ with distribution $\mu$ and $e^o \in \mathbb{R}^{N\times D_{o}}$ with distribution $\nu$. $\mathcal{T}_{\mu,\nu}= \{ T\in \mathbb{R}^{N,N}_{+}|T\mathds{1}_{N} = \mu, T^\top \mathds{1}_{N} = \nu\}$ represent the set of feasible transport plans. $T\in \mathbb{R}^{N, N}_{+}$ involving the marginal constraints of total mass equality between two distributions $\mu$ and $\nu$. The Gromov-Wasserstein Optimal Transport (GWOT) \cite{peyre2016gromov}, which leverages the relationships between intra-domain distances as its cost function, is deemed more appropriate for cross-modal alignment \cite{peyre2019computational,ryu2024cross}. Differently, we introduce constraints to limit transport operations within each class, aligning identical lesions across modalities. Specifically, we employed Labeled OT based on the Gromov-Wasserstein distance, ensuring that the optimization process remains consistent with GWOT. The additional constraints refine the $\mathcal{T}_{\mu,\nu}$ by incorporating both marginal constraints and class-specific restrictions, as defined below:
 \begin{equation}
     \mathcal{\bar{T}}_{\mu,\nu} = \{T\in \mathbb{R}^{N,N}_{+}| T \in \mathcal{T}_{\mu,\nu}, T_{ij}>0 \Rightarrow y^f_{i}=y^{o}_{j} \}.
 \end{equation}
Using from OCT to fundus alignment as an example, Labeled GWOT process aim to find an optimal transport plan $T^*\in\mathcal{\bar{T}}_{\mu,\nu} $ by sinkhorn iterations \cite{sinkhorn1967concerning}, the optimization process for the class-wise transport plan is defined as:
 \begin{equation}
     \min_{T_{c}\in \mathcal{\bar{T}}_{\mu,\nu}} \sum_{i,j,k,l}\underbrace{|(e^f_{i}-e^f_{k})^2-(e^o_{j}-e^o_{l})^2|^2}_{GW Distance} \times T_{c_{i,j}}T_{c_{k,l}} - \epsilon H(T_{c}),
    \label{eq1}
 \end{equation}
where $\epsilon H(T_{c})$ is the entropy regularization with the scaling factor $\epsilon$. Here, we define the optimal transport plan optimize by \autoref{eq1}: (1) from fundus to OCT class-wise transport as $T_c^{fo}$ and (2) from OCT to fundus class-wise transport plan as $T^{of}_{c}$. Then $T_c^{of}$ is used for OCT to fundus feature-wise transport and the optimization process for the feature-wise transport plan $T_v \in \mathbb{R}_{+}^{D_{o}\times D_{f}}$ is defined as:
\begin{equation}
         \min_{T_{v}} \sum_{i,j,k,l}\underbrace{|(e^f_{i}-e^f_{k})^2-(e^o_{j}-e^o_{l})^2|^2}_{GW Distance} \times T^{of}_{c_{i,j}}T_{v_{k,l}} - \epsilon H(T_v).
\end{equation}
Notably, as for the OCT-to-fundus alignment, we apply both class-wise and feature-wise transport. Meanwhile, for the fundus-to-OCT alignment, only class-wise transport is used. This distinction arises because fundus imaging offers a more comprehensive global perspective compared to the OCT data, making feature transport alignment more effective in capturing global structural information and semantic features within the fundus modality. Conversely, the OCT modality is subject to higher levels of noise, and the lesions are concentrated in localized regions. Therefore, relying primarily on class transport ensures the semantic alignment of corresponding lesions while avoiding additional noise. 
\subsection{OT-based Multi-Scale Alignment}

\noindent \textbf{Class-wise Alignment.} Here, we illustrate the fundus-to-OCT alignment as an example. Element $T^{fo}_{c_{i,j}}$ denotes the probability of transferring $e^f_{i}$ to $e^o_{j}$. And $T^{fo}_{c_{i,\cdot}}$ indicates the distribution over potential matches for the fundus representation $e^f_{i}$. 
We sample matching ground truth $e^o_{j}$ for $e^f_{i}$ according to the probability distribution of potential matches: $p(j|i) = \frac{T^{fo}_{c_{i,\cdot}}}{\sum_{j'} T^{fo}_{c{i,j'}}}$. As the training progresses over multiple epochs, the collection of all $e^o_{j}$ that is actually matched with $e^f_{i}$, can be interpreted in expectation as a `soft prototype' ($c^o_i$), which is defined as: 
\begin{equation}
    c^o_i = \mathbb{E}[e^o_{j}|e^f_{i}] = \sum p(j|i)e^o_{j}.
\label{eq5}
\end{equation}
In other words, each $e^f_{i}$ (or a batch of similar $e^f_{i}$) is associated with a weighted combination of possible matches that are determined through repeated random sampling. This stands in contrast to traditional clustering methods that typically employ a single (or a few) shared prototypes and strict one-to-one mappings. This design captures intra-class diversity while maintaining global consistency in the cross-modality alignment. We further train MLPs and treat each sample's class prototype as the ground truth to achieve cross-modal class prototype alignment. We use cosine similarity as the objective function for fundus to OCT ($\mathbb{L}_{\text{p\_OCT}}$) and OCT to fundus ($\mathbb{L}_{\text{p\_fundus}}$): 
\begin{align}
    &\mathbb{L}_{\text{p\_OCT}}=1- \sum_{i=1}^{n}\frac{e^{f}_{i}c^{o}_{i}}{\| e^{f}_{i}\|   \|c^{o}_{i}\|},
    &\mathbb{L}_{\text{p\_fundus}}=1- \sum_{i=1}^{n}\frac{e^{o}_{i}c^{f}_{i}}{\| e^{o}_{i}\|   \|c^{f}_{i}\|},
\label{eq6}
\end{align}
where primarily to promote semantic alignment in high-dimensional space rather than merely matching vector magnitudes.

\noindent \textbf{Feature-wise Alignment.} The feature level alignment is only used for OCT to fundus modality alignment. This is mainly due to the characteristics of OCT modality, that the lesion areas in OCT are concentrated and localized, while the fundus modality provides a global perspective, making such feature-wise global alignment only applicable to the fundus modality. During the alignment process from fundus to OCT, feature alignment is prone to overfitting to noisy patterns in OCT data, resulting in a decrease in the model's generalization performance. 
Given transport plan $T_v$, the OT fundus is identified by: $\text{OT fundus} = T_v^{\top} e^o$. OT fundus aligns the fundus feature distribution with OCT feature distribution, thereby preserving the latent structure across modalities and providing a global perspective across modalities in favour of the fundus modality.
\subsection{Asymmetric Fusion}
We designed asymmetric fusion to address the differences in characteristics between fundus and OCT modalities. For the fundus modality, we concatenate three types of features(the cross-modality feature from feature-wise alignment, the semantic consistency features produced by class-wise alignment and the feature from backbone) and feed them into an attention layer. The three types of features represent the correspondences between vascular structure and OCT stratification, lesion category priors, and local details representation, respectively. The attention layer adaptively learns nonlinear correlations between features through dynamic weight allocation, avoiding the inductive bias introduced by manually designed fusion rules \cite{vaswani2017attention}. For OCT modality, semantic consistency features after class-wise alignment and the feature extracted by the backbone are fed into MLPs. This design is based on the susceptibility of OCT images to motion artifacts and scattering noise interference, avoiding the overfitting of noise patterns in the training dataset by complex fusion modules. The two fused features are concatenated and fed into the classification head. The total loss function ($L_{total}$) integrates the classification loss and the alignment losses together:
\begin{equation}
    L_{total} = \mathbb{L}_{\text{ce}} +\mathbb{L}_{\text{p\_OCT}}+\mathbb{L}_{\text{p\_fundus}},
\end{equation}
where $\mathbb{L}_{ce}$ is a cross-entropy loss for the ophthalmic disease diagnosis task. The hyper-parameter each term's weight is set as 1 by default.
\section{Experiments}
\begin{table*}[t!]
\caption{Comparison with other methods. The performance of other models is from \textit{Liu et al.} \cite{liu2025Incomplete}, where our model follows the same experimental settings. The best results are \textbf{bolded}, with \underline{underline} indicating the second highest.}
\renewcommand\arraystretch{0.75}
\centering 
\setlength{\tabcolsep}{7pt} 
\scalebox{0.58}{
\begin{tabular}{@{\hskip 2pt}c@{\hskip 2pt}|@{\hskip 2pt}c@{\hskip 6pt}c@{\hskip 2pt}|ccc|ccc|ccc|ccc}
\toprule
\multirow{2}{*}{\textbf{Method}} & \multicolumn{2}{c|}{\textbf{Modality}} & \multicolumn{3}{c|}{\textbf{ Harvard-30k AMD}} & \multicolumn{3}{c|}{\textbf{Harvard-30k DR}}&\multicolumn{3}{c}{\textbf{Harvard-30k Glaucoma}}&\multicolumn{3}{|c}{\textbf{Average}}\\
\cmidrule(lr){2-3} \cmidrule(lr){4-6}\cmidrule(lr){7-9}\cmidrule(lr){10-12}\cmidrule(lr){13-15}

&\textbf{OCT} &\textbf{fundus}&\textbf{ACC}&\textbf{AUC}&\textbf{F1}&\textbf{ACC}&\textbf{AUC} &\textbf{F1}&\textbf{ACC}&\textbf{AUC}&\textbf{F1}&\textbf{ACC}&\textbf{AUC}&\textbf{F1}\\
\midrule
 \multicolumn{15}{c}{\textbf{Inter-Modality Missing with ResNet-50 Backbone}} \\
\midrule
2D-Resnet-50 & &\checkmark &73.12&75.35 &72.23&73.81&\underline{79.15}&70.47&73.12&75.35& 72.23&73.35&76.61&71.64\\
B-CNN $+$ distill & &\checkmark &72.35& 71.98& 70.03&73.62&67.50&69.68&73.39&76.61& 72.47&73.12&72.03&70.72\\
$\text{M}^{2}\text{LC}$ $+$ distill &&\checkmark&73.24&72.67&73.80&73.04&67.89&\underline{74.59}&72.78&70.23&71.11&73.02&70.26&73.16\\
IMDR  & & \checkmark &\underline{75.17}&\underline{80.48}&\textbf{76.59}&\underline{76.19}&79.07&72.18&\underline{75.54}&\underline{78.47}&\textbf{75.12}&\underline{75.63}&\underline{79.34}&\underline{74.63}\\
\rowcolor{gray!20} Ours & &\checkmark &\textbf{75.83}&\textbf{83.75}&\underline{74.46} &\textbf{82.74}&\textbf{85.39}& \textbf{82.70} &\textbf{75.72}&\textbf{78.81}&\underline{74.76}&\textbf{78.09}&\textbf{82.65}&\textbf{77.30}\\
\midrule
3D-Resnet-50 &\checkmark&& 65.17& 69.98& 69.63& 70.73 &69.94&61.97&65.72&69.89&\underline{70.98}&67.20&69.93&67.52\\
B-CNN $+$ distill & \checkmark&  &69.57& 70.14& 67.45&69.05& 65.25&67.93&69.64&68.95&67.18&69.42&68.11&67.52\\
$\text{M}^{2}\text{LC}$ $+$ distill  & \checkmark& &68.97&72.23&65.06&67.20&65.05&64.33&67.70&71.22& 65.60&67.95&69.50&64.99\\
IMDR  & \checkmark&  &\underline{70.62}& \underline{72.69}& \textbf{71.90}&\textbf{72.62}&\textbf{74.69}&\textbf{72.90}&\underline{71.16}&\underline{75.07}&70.37 &\underline{71.46}&\underline{74.15}&\underline{71.72}\\
\rowcolor{gray!20} Ours&\checkmark &&\textbf{74.50}&\textbf{79.21}&\underline{70.59}&\underline{72.02}&\underline{73.56}&\underline{71.81}
 &\textbf{75.47}&\textbf{80.94}& \textbf{75.13}&\textbf{73.99}&\textbf{77.90}&\textbf{72.51}\\
 \midrule
  \multicolumn{15}{c}{\textbf{Complete-Modality Fusion with ResNet-50 Backbone}} \\
 \midrule
B-CNN & \checkmark& \checkmark & 71.67&81.22&69.01&73.55&75.44&73.96&73.66&79.37 &72.89&72.96&78.67&71.95\\
B-EF  & \checkmark& \checkmark &69.33&79.12&60.64&74.98&74.13&77.17&74.19&77.52& 72.22&72.83&76.92&70.01\\
CR-AF & \checkmark& \checkmark & 73.00&81.37&67.73&76.57&78.35&76.27&73.92&77.48& 73.09&74.49&79.06&72.36\\
$\text{M}^{2}\text{LC}$  & \checkmark& \checkmark &74.00&\textbf{84.56}&70.73&77.36&83.62&76.01 &\underline{74.73}&77.45&73.54&75.36&\underline{81.87}&73.30\\
Eye-Most & \checkmark&\checkmark&\underline{75.55}&82.30&71.00&75.00&80.34&73.42&74.33&75.40& 72.01&74.96&79.34&72.08\\
IMDR & \checkmark& \checkmark & \textbf{76.45}&83.10&\underline{73.57}&\underline{78.03}&\underline{85.34}&\underline{78.23}&\textbf{76.50}&77.11& \textbf{74.36} &\underline{76.99}&81.85&\underline{75.38}\\
\rowcolor{gray!20} Ours& \checkmark& \checkmark &75.17&\underline{83.20}&\textbf{74.33}&\textbf{82.14} &\textbf{86.48}&\textbf{80.70}&74.39&\textbf{81.36} &\underline{73.72}&\textbf{77.23}&\textbf{83.68}&\textbf{76.25}\\
\bottomrule[1pt]
\end{tabular}
\label{tab:1}
}
\end{table*}
\textbf{Datasets and Evaluation.} We evaluate the proposed framework using three dataset: Harvard-30k AMD, Harvard-30k DR, and Harvard-30k Glaucoma from the publicly available multimodal datasets Harvard-30k \cite{luo2023eye}, covering Age-related Macular Degeneration (AMD), Diabetic Retinopathy (DR), and
Glaucoma diseases. Harvard-30k subsets are annotated with four-tier AMD grading and two-tier for glaucoma and DR, with dimensions of 448 $\times$ 448 for fundus and 200 $\times$ 256 $\times$ 256 for OCT. To ensure reliable results, each dataset underwent five-fold cross-validation. Performance was evaluated using Accuracy (ACC), F1-Score (F1), Area Under the Curve (AUC). 

\noindent \textbf{Implementation Details.} All compared models use ResNet50 \cite{he2016deep} as the backbone for both OCT and fundus modality. The input fundus images were resized to 384 $\times$ 384; OCT images were resized to 96 $\times$ 96 $\times$ 96.  All
of the models were trained with a batch size of 32 for the same number of epochs. 

\noindent \textbf{Experiments Setting.} To comprehensively evaluate the performance of the proposed models, we designed the following three experimental setups: 1) \textbf{Complete-Modality Fusion:} Train on the complete training set and evaluate on the complete testing set; 2) \textbf{Inter-Modality Missing:} Train on a complete training set and evaluate on test set with single modality (OCT or fundus); 3) \textbf{Proportional Random Missing:} Train on the training set with a partial proportion of missing modalities and evaluate on the testing set with a partial proportion of missing modalities (same missing ratios for training and test sets).

\begin{figure}[t!]
    \centering
    \includegraphics[width=1\linewidth]{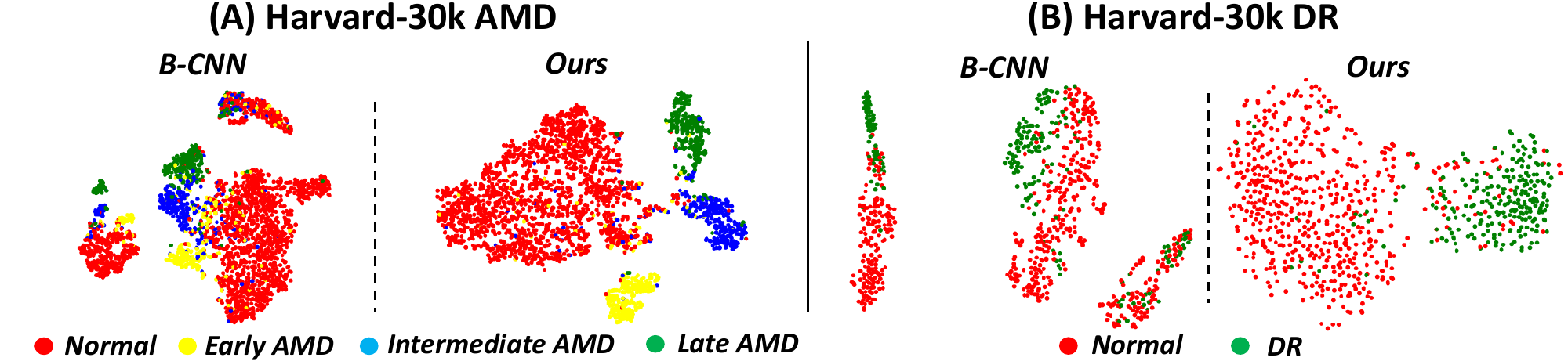}
    \caption{A t-SNE visualization of the feature distribution from B-CNN and our model, captured before the final FC layer on the AMD and DR datasets.}
    \label{tsne}
\end{figure}

\noindent \textbf{Complete-Modality Fusion.} we compare our model with six representative complete-modality fusion methods, including B-CNN(baseline model utilizing an intermediate multimodal fusion method), B-EF \cite{hua2020convolutional}, CR-AF \cite{zheng2023casf}, $\text{M}^2$LC \cite{woo2018cbam}, Eye-Most \cite{zou2024confidence}, IMDR \cite{liu2025Incomplete}. As shown in \autoref{tab:1}, our model consistently outperforms other methods across all datasets and backbones evaluated. For instance, our model outperforms Eye-most \cite{zou2024confidence} 5.8\% on F1 and 5.4\% on AUC on average and exceeds IMDR \cite{liu2025Incomplete} 1.1\% on F1 and 2.2\% AUC on average. To illustrate the effects of our model, we perform a t-SNE visualization (shown in \autoref{tsne}) on AMD and DR testset under Complete-Modality Fusion.
\begin{figure}[t!]
    \centering
\includegraphics[width=0.9\linewidth]{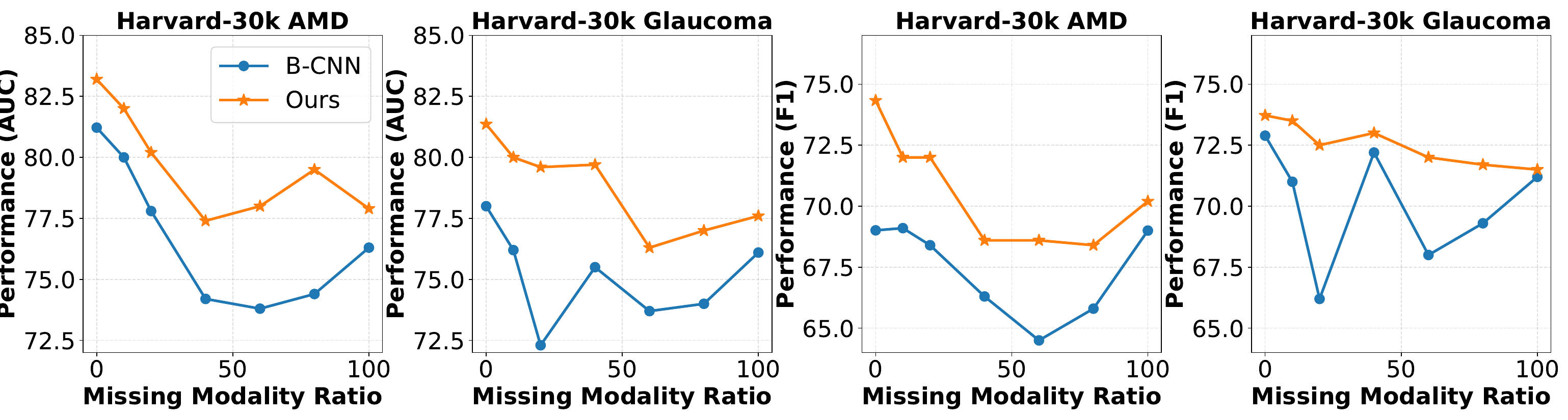}
    \caption{Comparison of B-CNN and Ours in proportional random missing experiment, evaluated across various missing rates on different ophthalmic diseases and datasets.}
    \label{fig:miss}
\end{figure}

\noindent \textbf{Inter-Modality Missing.} As shown in \autoref{tab:1}, 
Compared to the evaluation strategy with complete modality fusion, the performance of all other models declines across different datasets when modality is missing. In contrast, our model maintains outstanding performance, exceeding IMDR \cite{liu2025Incomplete} 4\% AUC onaverage when OCT is missing and  5\% when fundus is missing. Notably, while the OCT data in the AMD and Glaucoma datasets manifests highly distinct lesion characteristics that are readily discernible by medical professionals, those in the DR dataset are relatively inconspicuous. 
The diagnostic performance of our model in the event of fundus modality absence is highly congruent with the aforementioned phenomena. This unique trait, however, remains unobserved in other models. This compellingly attests to the fact that our model can precisely and effectively align the semantic features of lesions.

\noindent \textbf{Proportional Random Missing.}
As shown in \autoref{fig:miss}, by randomly masking a portion of the modality data from 0\% to 100\%, we compared our model with B-CNN in the AMD and Glaucoma datasets. Our model performs outstandingly and demonstrates extremely strong robustness, maintaining relatively stable performance compared to the baseline models. This indicates that our proposed model effectively aligns the key lesion features and reconstructs information under missing modalities setting, equipping it to handle interferences in real-world scenarios.

\begin{table*}[t!]
\caption{Ablation study of each components on Harvard-30k DR dataset.}
\renewcommand\arraystretch{0.75}
\centering
\setlength{\tabcolsep}{7pt} 
\scalebox{0.75}{%
\begin{tabular}{c|cc|ccc}
\hline
Variants&OCT&fundus&ACC& AUC  & F1  \\
\hline
\textit{w/o} Class-wise Alignment&&\checkmark&77.38&76.20&81.42\\
\rowcolor{gray!20} \textbf{Ours}&&\checkmark&\textbf{82.74}&\textbf{85.39}& \textbf{82.70}\\
\hline
\textit{w/o} Class-wise Alignment &\checkmark&&70.24&68.14&70.88\\
\textit{w/o} Feature-wise Alignment&\checkmark&&70.83& 73.11&70.69\\
\rowcolor{gray!20} \textbf{Ours}&\checkmark&&\textbf{72.02}&\textbf{73.56}&\textbf{71.81}\\
\hline
\textit{w/o} Asymmetric Fusion&\checkmark&\checkmark&79.17& 85.42&77.80\\
\rowcolor{gray!20} \textbf{Ours}&\checkmark&\checkmark&\textbf{82.14}&\textbf{ 86.48}& \textbf{80.70} \\
\hline
\end{tabular}%
}
\label{ab_data_pre}
\end{table*}
\noindent \textbf{Ablation Study of Proposed Components.}
To assess the effectiveness of
our model, we conducted a thorough ablation study on the DR dataset, shown in \autoref{ab_data_pre}. We evaluated the performance of the model by removing the Class-wise Alignment, Feature-wise Alignment, and Asymmetric Fusion modules individually and remaining the rest of the model structure unchanged. Notably, removing the Class-wise Alignment has the most significant impact on AUC (7\% for fundus modality, 8\% for OCT modality). Feature-wise Alignment and Asymmetric Fusion modules also contribute to the improvement of the DR grading effect, such as 0.6\% and 1.2\% of AUC. These experimental results demonstrate the effectiveness of each component based on different modality properties.

\section{Conclusion}
We proposed  a novel multimodal alignment and fusion framework. Our model, based on Labeled OT, has achieved  multi-scale alignment at both the class and feature levels. This ensures precise matching of key lesion features to obtain robust and effective multimodal feature representations while maintaining excellent performance even in the presence of missing modalities. Furthermore, we fully consider the characteristics of different modalities and design an asymmetric fusion mechanism to maximize the advantages of each modality. Experimental results on three datasets demonstrate that our method significantly outperforms state-of-the-art models under different experimental settings. In particular, our approach has exhibited exceptional robustness to missing modalities, further validating its effectiveness across different ratio of modality missing in real-world scenarios.
\begin{credits}
\subsubsection{Acknowledgments}
This work is supported by Y. Meng's The Royal Society Fund (IEC\textbackslash
NSFC\textbackslash
242172), United Kingdom. This work is also partly funded by the UK Engineering and Physical Sciences Research Council [grant number EP/X01441X/1] and Medical Research Council (grant number MR/Z506175/1). Qinkai Yu thanks the PhD studentship funded by Liverpool Centre for Cardiovascular Science and University of Exeter.
\subsubsection{\discintname}
The authors have no competing interests to declare that
are relevant to the content of this article.
\end{credits}

%

%
%
\bibliographystyle{splncs04}
\bibliography{paper-0878}
%




\end{document}